\documentclass{article}

\usepackage[utf8]{inputenc} 
\usepackage{hyperref}       
\usepackage{url}            
\usepackage{booktabs}       
\usepackage{amsfonts}       
\usepackage{nicefrac}       
\usepackage{microtype}      
\usepackage{lipsum}
\usepackage{graphicx}
\graphicspath{ {./images/} }
\usepackage{algorithm}
\usepackage{algorithmic}
\usepackage{enumerate}
\usepackage{amssymb}
\usepackage{amsmath}
\usepackage{mathrsfs}
\usepackage{tabularx}
\usepackage{comment}
\usepackage{delarray}
\usepackage{times}
\usepackage{subfigure}
\usepackage{color}
\usepackage{url}
\usepackage{fullpage}

\usepackage{helvet} 
\usepackage{courier}
\usepackage{caption}
\title{Mitigating Bias in Federated Learning}

\author{Annie Abay, Yi Zhou, Nathalie Baracaldo, Shashank Rajamoni, Ebube Chuba, Heiko Ludwig}
\date{IBM Research}

\begin{document}
\maketitle
\begin{abstract}
As methods to create discrimination-aware models develop, they focus on centralized ML, leaving federated learning (FL) unexplored. FL is a rising approach for collaborative ML, in which an aggregator orchestrates multiple parties to train a global model without sharing their training data. In this paper, we discuss causes of bias in FL and propose three pre-processing and in-processing methods to mitigate bias, without compromising data privacy, a key FL requirement.
As data heterogeneity among parties is one of the challenging characteristics of FL, we conduct experiments over several data distributions to analyze their effects on model performance, fairness metrics, and bias learning patterns.
We conduct a comprehensive analysis of our proposed techniques, the results demonstrating that these methods are effective even when parties have skewed data distributions or as little as 20\% of parties employ the methods.
\end{abstract}

\section{Introduction}
As machine learning (ML) has been applied to facilitate decision-making in various areas, such as hiring, loan grading etc., there have been and continue to be increasing concerns \cite{Bhandari2016Big, sweeney2013discrimination} that ML models will inevitably ``learn undesired bias'' from the training data and make unfair predictions. 
From algorithms erroneously detecting suspects of crimes base on the color of their skin \cite{naturebias}
and deciding who goes to jail \cite{mlarrests},
to algorithms used to predict test scores that provide unfairly higher scores to socio-economically privileged students, allowing them to enter universities at a higher rate \cite{rahim2020uk},
the lack of understanding and control of undesired bias in ML models has tangible consequences.
To deal with this challenge of biased models, many researchers have devoted their efforts, e.g., \cite{kamiran2012rw,kamishima2012,zafar2015fairness,dwork2012fairness, bellamy2018ai} to define, detect and mitigate bias in ML over the past decade.
These approaches mainly measure and reduce undesired bias with respect to a sensitive attribute, such as \textit{age} or \textit{race}, in the training dataset. 
Although many of them provide various effective approaches, they all focus on centralized ML, where the training dataset is stored in a single place, executing the learning procedure, and hence assume full access to the entire dataset.

This assumption is not true for \textit{federated learning} (FL) \cite{mcmahan2016communication, konevcny2016federated}, where data is not shared with a central entity for reasons of privacy, 
confidentiality or regulatory constraints, such as the European General Data Protection Regulation (GDPR), California Consumer Privacy Act (CCPA), or the Health Insurance Portability and Accountability Act (HIPAA).
In FL, multiple parties collaboratively train a model without sharing their raw data; they share \textit{model updates}, such as model parameters or gradients of their local models, with the aggregator during each round of training; the aggregator uses these to update the global model. The aggregation of the \textit{model updates} usually follows a specific \textit{fusion} strategy. 
This process repeats for several rounds until a certain accuracy or maximum number of training rounds has been reached.

Finding methods to both measure and reduce bias without directly examining sensitive information conflicts with the full data access most current techniques require, and is an open issue in the field \cite{book-challenges-fl}.

Consider a scenario where a bank in the European Union wants to train a ML model to determine whether a potential client is qualified to obtain a loan. Due to legislation in different countries, it is not possible to transmit all data from other branches to a central place \cite{turkey-legislation}. The bank thus decides to engage all of its country's branches in an FL process to train a model without transferring data. Each branch may have a slightly different data distribution, or number of samples, based on its location. A branch in an underdeveloped town, or near a women's college, or surrounded by multiple luxury apartments is likely to disproportionately represent a specific subgroup, making it difficult to identify and mitigate bias on local datasets only. 
Even if the data distribution between branches is similar, bias in the final model can still exist, which is demonstrated in our experiments.
In this scenario, avoiding bias caused by sex or race is relevant.

Current state-of-the-art techniques were not designed to handle several difficulties that emerge in FL settings.
In particular, parties may have heterogeneity in distribution and amounts of data that, due to privacy concerns, cannot be freely shared. 
These requirements have led to the design of solutions that apply multi-party computation during the training process \cite{truex2019hybrid,xu2019hybridalpha} or add differential privacy noise \cite{shokri2015privacy}.
Under these circumstances, bias detection techniques that assume the training set distribution is known, e.g. {\cite{kamiran2012decision}}, cannot be applied. The aggregator and parties have a partial view of the entire data, and so it is difficult to measure how different model updates are affecting the global model during training. This lack of visibility is exacerbated by the dynamic participation of parties, which may drop out of the system due to connectivity constraints for a few training rounds.

In this paper, we address these important gaps.
We present an analysis of the sources of bias in FL settings and adapt two highly utilized fairness techniques, resulting in the proposal of \textit{local reweighing}, \textit{global reweighing with differential privacy} and \textit{federated prejudice removal}.
We further assess these approaches under several FL settings, including multiple data distributions when sensitive attributes are known.
Additionally, we assess the effectiveness of four relevant fairness metrics that determine how biased a model is, and found one of them unreliable in FL. We summarize the \textbf{contributions} of this paper as follows:

\begin{itemize}
    \item\textbf{Party-based Bias Mitigation:} 
    We propose three bias mitigation techniques based on 
    well-known centralized bias mitigation techniques to train models via FL without compromising data privacy and prediction accuracy. 
    \item\textbf{Fairness and Privacy Trade-off:} 
    We propose three approaches with different privacy designs. 
    We analyze their trade-offs between model performance, privacy and simplicity.    
    \item \textbf{Uncooperative Parties:} 
    We study the effect of parties refusing to get involved in bias mitigation procedures and we provide a bias mitigation technique that can ensure the final model is fair even under such circumstances. 
    \item \textbf{Highly Imbalanced Data Distribution:}
    We demonstrate the performance of our proposed methods under a variety of data heterogeneity settings in terms of skewed datasets and several sample ratios for multiple sensitive attributes, even extreme cases.
    \item \textbf{Stability of fairness metrics in FL:} We systematically assess popular bias metrics in FL setting and demonstrate some of them are not suitable.  
\end{itemize}

This paper is organized as follows:

We discuss related work in Section~\ref{sec:related}.  
Section \ref{sec:causes} and \ref{sec:main} are devoted to discussing the causes of bias and our approaches to mitigate some of them in FL. 
Extensive experiments are conducted in Section \ref{sec:exp} to demonstrate the effectiveness of our proposed methods.
Section \ref{sec:end} includes our concluding remarks.

\noindent{\bf Terminology and notation:}

Throughout the paper, 
\textit{favorable}, e.g., $1$ and \textit{unfavorable labels}, e.g., $0$, refer to advantageous and disadvantageous outcomes. \textit{Protected/sensitive attributes} are features that divide a group and may impact treatment, i.e. sex, race, age, etc. \textit{Privileged} and \textit{unprivileged groups} are based around sensitive attributes, and are understood to benefit from and suffer from bias, respectively.  
$D := (X, Y)$ denotes the training dataset, where $X$ refers to the feature set and $Y$ refers to the label set. $S\subseteq X$ and $s/s_i$ are the set of sensitive attributes or a specific sensitive attribute value, respectively. 
Similarly, $x/x_i$ and $y$ denote a feature vector and label.

\section{Related work}\label{sec:related}
In the past decade, there has been fruitful research addressing the concern of producing biased ML models. 
For a given sensitive attribute, existing methods try to ensure that the main factor in determining the outcome of the model, positive or negative, is not the sensitive attribute itself.

They can be classified depending on the stage of the training process they are incorporated into, resulting in three categories: \textit{pre-processing}, \textit{in-processing} and
\textit{post-processing} techniques.
Pre-processing methods, e.g., \cite{kamiran2012rw, feldman2015certifying, dwork2012fairness,zemel2013learning}, work to create a less biased dataset, and they either modify the raw data samples by changing the value of sensitive attributes or class labels, or assign sample weights to data samples.
All pre-processing techniques can be applied regardless of the chosen ML models and training algorithms, but may lead to unstable model performance and losses in prediction accuracy.
In-processing methods \cite{kamishima2012, calders2010three, goh2016satisfying, zafar2015fairness} usually modify the optimization problem associated with the chosen classifiers by adding either a discrimination-aware regularizer to the objective function or bias mitigation constraints to the optimization formulation.
A major drawback of this type of approach is that it is tied to a specific ML model and training algorithm, e.g., \cite{kamishima2012} only works for logistic regression models. 
Post-processing methods \cite{hardt2016equality, pleiss2017fairness,kamiran2012decision} are used to help trained classifiers make fairer predictions for a provided test dataset.
All of the aforementioned studies focus on centralized ML, so cannot be applied as is to a FL setting due to the raw data sharing.

Only a few recent results \cite{mohri2019agnostic,hsu2020fedvision,liang2020think} cover fairness in FL settings, but only discuss development of fair FL models in terms of contribution fairness, i.e., making sure the global model learns from parties equally, but not fairness related to sensitive attributes. None have measured the resulting models' fairness performance against well-known fairness metrics.

\section{Fairness Analysis in FL: Causes }\label{sec:causes}

We now analyze possible causes of bias in FL: 

\begin{table*}[t!]
\caption{\footnotesize Summary of our proposed approaches}
\label{tab:comparison}
\begin{center}
\begin{small}
\begin{sc}
\begin{tabular}{cccc}
\hline
\textbf{Methods}                                    & \textbf{Privacy}   & \textbf{\begin{tabular}[c]{@{}c@{}} Additional\\ Communication\end{tabular}} & \textbf{Hyperparameters} \\ \hline
{Local Reweighing}                                                                         &
Same as plain FL
& 0                                                                           & none          \\ \hline
{ \begin{tabular}[c]{@{}c@{}}Global Reweighing\\ with Differential Privacy\end{tabular}} & $\epsilon$-Differential privacy                              & 1.5                                                                                 & $\epsilon$  \\ \hline \begin{tabular}[c]{@{}c@{}}Federated \\  Prejudice Removal \end{tabular} &
Same as plain FL  
& 0                                                                                    & $\eta$      \\ \hline
\end{tabular}
\end{sc}
\end{small}
\end{center}

\end{table*}
\noindent\textit{Traditional bias sources:}
From an individual party's perspective, local training in FL is similar to training a model in a centralized ML.
Sources of unfairness \cite{kamishima2012}, like {\sl prejudice}, {\sl underestimation}, and {\sl negative legacy}, which have been previously identified in centralized ML, are also underlying causes of bias in FL settings.
With multiple training sets involved in FL, each party will introduce its own biases to the global model via the shared model updates.
Due to the interactions between the parties and the aggregator during the training process, the following factors also influence the fairness of the final model.

\noindent \textit{Party selection, sub-sampling and drop outs:}
Existing FL approaches such as {\cite{mcmahan2016communication,chai2020tifl}} do not query all parties equally.
Whether someone gets to participate in one round of FL training may be correlated with sensitive attributes and creates undesired bias. 
If a company uses cell phone data to train a FL model, factors such as network speed can impact whether a user’s data is collected, which is linked to socioeconomic status. Many FL algorithms decide whether or not to query parties and hence check if a cell phone is charged or plugged in to decide, which could correlate with factors like day-shift and night-shift work schedules \cite{book-challenges-fl}.

\noindent \textit{Data heterogeneity:} 
Even when all parties are queried, another challenging but under-investigated aspect of FL is that the underlying data of each party may differ. As mentioned in our banking example, a branch nearby a women's college would result in a data comprising mostly of women, which is likely quite different than the overall composition of the bank's customer dataset.

\noindent\textit{Fusion algorithms:}
The fusion algorithms used by the aggregator to combine the parties' model updates may induce bias based on whether the aggregator performs an equal or weighted average.
FL algorithms may weigh higher the contributions from populations with more data \cite{mcmahan2016communication}, i.e., heavy users of specific products, amplifying effects of over-/under-representing specific groups in a dataset.

There are two main use cases in FL: \textit{i)} when parties are cell phones and IoT devices and usually, the number of parties is vast and
\textit{ii)} when parties are companies, subsidiaries or data centers, and typically, there are fewer parties. In this paper, we address the second case, which has started to be specially important given that large companies have started to offer products in this line \cite{ibmfl,nvidiaclara} to train ML models in healthcare, banking \cite{banking}, and insurance \cite{banking}.

\section{Proposed Approaches}\label{sec:main}
In this section we propose three approaches to mitigate bias in FL using two different privacy designs: \textit{i)} avoiding transfer of sensitive information and \textit{ii)} adding differential privacy as shown in Table \ref{tab:comparison}.
Our approaches are designed based on two centralized bias mitigation techniques, \textit{Reweighing} and \textit{Prejudice Remover}, both detailed below:

\subsection{Reweighing for FL}\label{rw-fl}
Reweighing \cite{kamiran2012rw} is a centralized pre-processing bias mitigation method, which works primarily by attaching weights to samples in the training dataset. 
It accesses the entire training dataset and computes weights as the ratio of the expected probability ($P_{exp}$) to the observed probability ($P_{obs}$) of the sample 's sensitive attribute/label pairing (see equation \eqref{eq:reweigh}).

As FL has become popular due to increasing data privacy concerns, mitigating biases in FL cannot come along with the price of losing privacy protection. 
With this in mind, we propose two ways to adapt the Reweighing method in FL settings.

\noindent\textbf{Local reweighing:} To fully protect parties' data privacy, each party computes reweighing weights locally, based on its own training dataset, during pre-processing and then uses the reweighing dataset for its local training. 
Therefore, parties do not need to communicate with the aggregator or reveal their sensitive attributes and data sample information.
We will demonstrate in Section~\ref{sec:exp} that \textit{local reweighing} is effective without compromising the prediction accuracy, even when only a subset of the parties employ it or parties have highly skewed local data distribution.

\begin{algorithm}[ht]
   \caption{Global Reweighing with DP}
   \label{alg:localrew}
\begin{algorithmic}[1]
   \STATE {\bf Input:} Differentially private parameter $\epsilon$.
    \STATE The aggregator queries parties in $\mathcal{P}$ for their counts per sensitive attribute/label pairing.
    \STATE Aggregator queries all parties for noisy counts  
    \STATE each party $p_i$ computes:
    \begin{align}\label{eq:reweigh1}
         C_i(s,y,\epsilon) &:=|(X \in D_i | S = s) \wedge (Y = y)|  \nonumber\\
         &\qquad  + \mbox{DP\_noise}(\epsilon), \forall s\in S, y\in Y.
    \end{align}
    \STATE The aggregator computes
        \begin{align}\label{eq:reweigh}
        W(s,y) := \tfrac{P_{exp}(s,y)}{P_{obs}(s,y)} = 
        \tfrac{\sum_{i, y\in Y}C_i(s,y,\epsilon)*\sum_{i, s\in S}C_i(s,y,\epsilon)}{C_i(s,y,\epsilon)  \sum_{i, s\in S, y\in Y}C_i(s,y,\epsilon)},
        \end{align}
    \STATE Aggregator shares $W(s,y)$ with all parties in $\mathcal{P}$.
    \STATE Each party $p_i$ initializes $D_{i,w} = \emptyset$ and assigns $W(s,y)$ to data entry $(x,y)$ as follows: 
    \FOR{$(x,y)\in D_i$ with sensitive attribute $s$}
    \STATE Add data entry $(x, y, W(s, y))$ to $D_{i,w}$.
    \ENDFOR
    \STATE Each party uses pre-process data $D_{i,w}$ for training.
\end{algorithmic}{}
\end{algorithm}

\noindent\textbf{Global reweighing with differential privacy (DP):} If parties agree to reveal their sensitive attributes and
their noisy sample counts with the aggregator, a \textit{differentially private global reweighing} approach can be applied.

In particular, during the pre-processing phase, the aggregator will collect statistics such as the noisy number of samples with privileged attribute values and favorable labels from parties, compute \textit{global reweighing} weights $W(s,y), \forall s, y$ based on the collected statistics, and share them with parties.
Parties assign the \textit{global reweighing} weights to their data samples during FL training.
As shown in \eqref{eq:reweigh1}, DP-noise is injected according to well-known privacy mechanisms. We used \cite{naoise:library}.
By adjusting the amount of noise injected via adjusting $\epsilon$, parties can control their data leakage in the pre-processing phase, while still mitigating bias via the \textit{global reweighing} method.
Our method adds noise to the counts of our sensitive-label value pairings, thereby adjusting the weight calculation.
One limitation of \textit{global reweighing} is that it does not apply to FL systems with dynamic participation,
as the global reweighing weights would re-calibrate in relation to the number and size of training sets changing over the course of training.

\noindent{\bf Remark}: The proposed reweighing methods for FL are conducted only during pre-processing phase and try to mitigate bias by adjusting sample weights and hence training data distribution. Therefore, these two methods \textit{will not} affect the convergence behavior of FL algorithms used to train the model, like FedAvg\cite{mcmahan2016communication}, etc.

\subsection{Prejudice Remover for FL}\label{pr-fl}
Prejudice Remover \cite{kamishima2012} is an in-processing bias mitigation method proposed for centralized ML, which works by adding a fairness-aware regularizer, $R(D,\Theta)$, to the regular logistic loss function as follows:
\[
\mathcal{L}(D;\Theta) + \tfrac{\lambda}{2}||\Theta||_{2}^2+ \eta R(D,\Theta).
\]

Here $\mathcal{L}$ denotes a regular loss function, $\| \Theta\|_{2}^2$ is a $\ell_2$ regularizer that protects against over-fitting, $R$ represents the regularizer which penalizes biased classifiers, $\Theta$ represents the model parameters, and
$\lambda$ and $\eta$ are regularization parameters.
$R$ aims to reduce the prejudice index \cite{kamishima2012}, which measures the learned prejudice from the training dataset, and is defined as follows:
\[
R(D,\Theta) = \sum_{(x_{i},s_{i}) \in D} \sum_{y \in {0,1}} M[y| x_{i},s_{i}; \Theta] \ln{\tfrac{\hat{P}r[y|s_{i}]}{\hat{P}r[y]}},
\]

where $M[y| x_{i}, s_{i}; \Theta]$ denotes the conditional probability of prediction equals $y$ given a data sample with non-sensitive features $x_{i}$, sensitive attribute $s_{i}$ and $\Theta$, and $\hat{P}r$ is the sample distribution induced by the training dataset.

From second equation, we can see that evaluating $R$ requires knowledge of local data distribution, and may result in data leakage if the evaluation is performed globally.

We propose \textbf{federated prejudice removal}, in which each party applies the Prejudice Remover algorithm \cite{kamishima2012} to train a less biased local model, and shares only the model parameters with the aggregator. The aggregator can then employ existing FL algorithms, like simple average and {\sl FedAvg} \cite{mcmahan2016communication}, etc., to update the global model. 
Besides parties not revealing sensitive attributes, another advantage of \textit{federated prejudice removal} is that it only modifies the party's local training algorithm, and leaves the fusion of model parameters untouched. 
In this case, our proposed approach can be integrated with a variety of existing FL algorithms, even robust aggregation methods like Krum \cite{blanchard2017machine} and coordinate median \cite{yin2018byzantine} to defend against adversarial attacks. 
It is important to note that Prejudice Remover is just one of the in-processing methods that mitigate bias via adding a regularizer term to the objective function or enforcing some fairness-aware constraints; similar approaches can also be found in \cite{kamishima2012, calders2010three, goh2016satisfying, zafar2015fairness}.
Similar to \textit{federated prejudice removal}, these types of in-processing methods can also be extended to a FL setting following the proposed strategy.

The major disadvantage of \textit{federated prejudice removal} is selecting a reasonable coefficient $\eta$ for the regularizer $R$. The value of $\eta$ controls the trade-off between the prediction accuracy and fairness during each round of a party's local training. 

As $\eta$ increases, parties mitigate more bias and move toward a fairer local model, but also experience worsening performance metrics, specifically accuracy \cite{kamishima2012}. 
Tuning $\eta$ can be extremely tedious in a FL system since parties may have different sizes of training datasets and local data distributions, and the performance of
parties' local models have an indirect impact on the final global model's performance depending on the specific FL algorithm.

\section{Experimental Results}\label{sec:exp}
We now conduct a comprehensive set of experiments to examine the aforementioned mitigation methods in FL settings. 

First, we examine the case where all parties in the FL setting have similar data distributions (independently and identically distributed (IID)).
We then examine real-world scenarios where parties' training datasets have disproportionate privileged/unprivileged ratios. These experiments allow us to see how effective our proposals are under a variety of circumstances.

\noindent{\bf Experimental setup:}
For all experiments presented in the following subsections, we trained a $\ell_2$- regularized logistic regression model in a FL fashion.
We utilize the UCI Adult Dataset \cite{uciadult} 
and the ProPublica Compas (Recidivism) Dataset \cite{ucicompas}, two standard datasets used in the fairness literature, e.g., \cite{bellamy2018ai,zafar2015fairness,hardt2016equality}.

\noindent \textit{Adult dataset:}
The Adult dataset contains $48,842$ samples and classifies whether individuals make more or less than $50$k per year, based on census data. 
The sensitive attributes we evaluate are \textit{sex} and \textit{race}, whose values are mapped to 0 or 1, for the unprivileged and privileged groups respectively.  In this case, White: 1, Black: 0, Male: 1, Female: 0. Class values are mapped to 0 ($<=$50K) and 1 ($>$50K) for the negative and positive classes respectively \cite{zafar2015fairness}.
We preprocess the data prior to splitting amongst parties. 
All features except sex, race, age, education and class were dropped. We binned the feature values of \textit{age} and \textit{education} using intervals of 10. 

\noindent \textit{Compas dataset:}
The Compas dataset contains $7,215$ samples and classifies whether individuals who have broken the law in the past two years will reoffend. 
The sensitive attributes we evaluate are \textit{sex} and \textit{race}, whose values are also mapped to 0 or 1 for the unprivileged and privileged groups respectively. In this case, White: 1, Black: 0, Male: 0, Female: 1. Class values are mapped to 0 (will break law again) and 1 (will not break law again) for the negative and positive classes respectively \cite{zafar2015fairness}.
We again preprocess the data prior to splitting amongst parties.
All features except sex, race, age, prior count, charge degree and class were dropped; before this, samples with a charge date outside of 30 days from when they were arrested are filtered out, as well as samples that did not receive jail time. 
The values of \textit{age} are binned into categories of Under 25, 25-45, and Over 45; \textit{prior count} values are binned into categories of 0, 1-3, and $>$3. %

Similar to the \textit{Adult} setting, we evaluate sensitive attributes \textit{sex} and \textit{race} for \textit{Compas}. In this case, however, the unprivileged and privileged groups for the \textit{sex} attribute are flipped; female is privileged and male is unprivileged. The sensitive attributes we evaluate are \textit{sex} and \textit{race}. To ensure that the reported results are not skewed due to the variations in different testing sets, we utilize a stratified \textit{global testing set}, sampling 20\% of the data from the original dataset.

\textit{Evaluation of fairness metrics:}
We will evaluate all resulting models via four popular fairness metrics: \textit{statistical parity difference}, \textit{equal opportunity odds}, \textit{average odds difference} and \textit{disparate impact}. These metrics are calculated via different aspects of the confusion matrix.
In particular,
\textit{statistical parity difference} and \textit{disparate impact} are the ratio and difference, respectively, of the success rate between the unprivileged and privileged groups. \textit{Equal opportunity difference} is the true positive rate difference between the unprivileged and privileged groups, and \textit{average odds difference} is the mean of the false positive rate difference and the true positive rate difference, both between the unprivileged and privileged groups.

All FL experiment results are compared against the centralized version as a baseline.

We analyze the resulting ML models from two aspects: prediction performance and fairness performance. 
For prediction performance, we use accuracy and F1 score.
As fairness is a complex concept, there is no single best metric that measures all aspects of it \cite{bellamy2018ai}, which in turn asks for multiple metrics for evaluation. 

\textit{Evaluation of fairness metrics:}
We will evaluate all resulting models via four popular fairness metrics: \textit{statistical parity difference}, \textit{equal opportunity odds}, \textit{average odds difference} and \textit{disparate impact}. These metrics are calculated via different aspects of the confusion matrix.
In particular,
\textit{statistical parity difference} and \textit{disparate impact} are the ratio and difference, respectively, of the success rate between the unprivileged and privileged groups. \textit{Equal opportunity difference} is the true positive rate difference between the unprivileged and privileged groups, and \textit{average odds difference} is the mean of the false positive rate difference and the true positive rate difference, both between the unprivileged and privileged groups.
The ideal value for \textit{SPD}, \textit{EOD} and \textit{AOD} is 0, and for \textit{DI} is 1.
 
All results are averaged over three runs for each experimental setup.

\subsection{Performance under similar data distributions (IID)}\label{sec:exp-stratified}

We now study how our proposed approaches behave under FL systems with IID data. 
We examine how different factors in FL, apart from data distribution, can affect the global model's fairness performance.

The aim of this set of experiments is to examine how different factors in FL apart from party data distribution can affect the global model's fairness performance.
We split the \textit{Adult} training set in a stratified fashion amongst eight parties of equal size, resulting in $4,884$ samples each; the same is done for the \textit{Compas} dataset amongst five parties, resulting in $1,154$ samples each.

\begin{figure}[h]
\centering
\includegraphics[width=0.7\columnwidth]{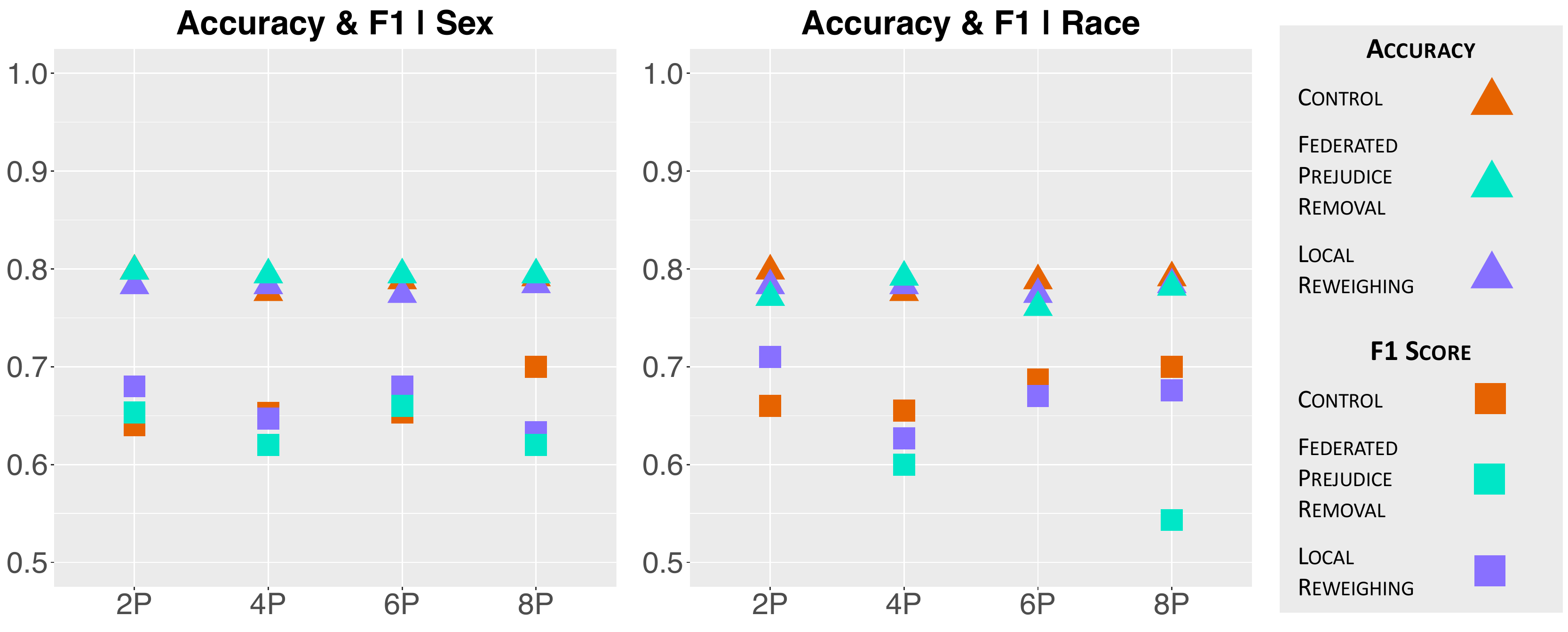}
\caption{\footnotesize Performance metrics of models trained on stratified \textit{Adult} dataset}
\label{graph:adult-stratified}
\end{figure}

\begin{figure}[h]
\centering
\includegraphics[width=0.7\columnwidth]{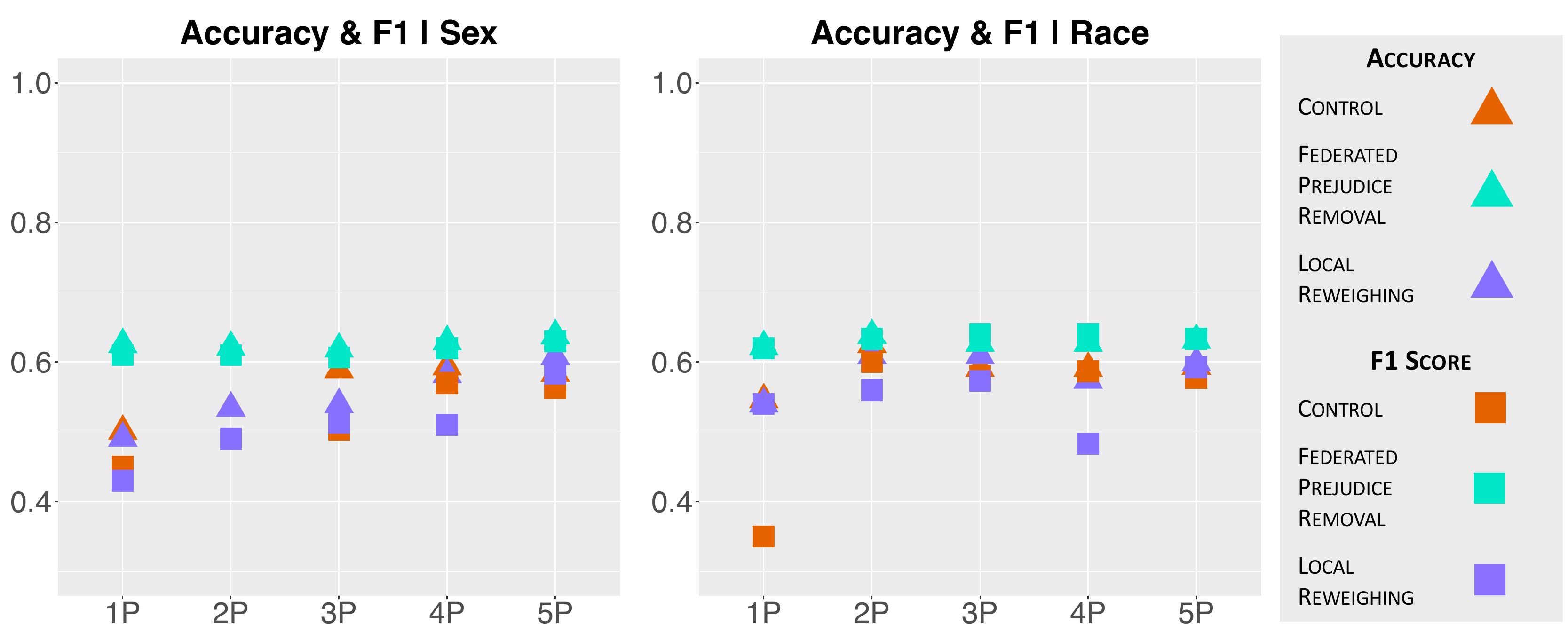}
\caption{\footnotesize Performance metrics of models trained on stratified \textit{Compas} dataset}
\label{graph:compas-acc-stratified}
\end{figure}

\begin{figure}[h]
\includegraphics[width=\columnwidth]{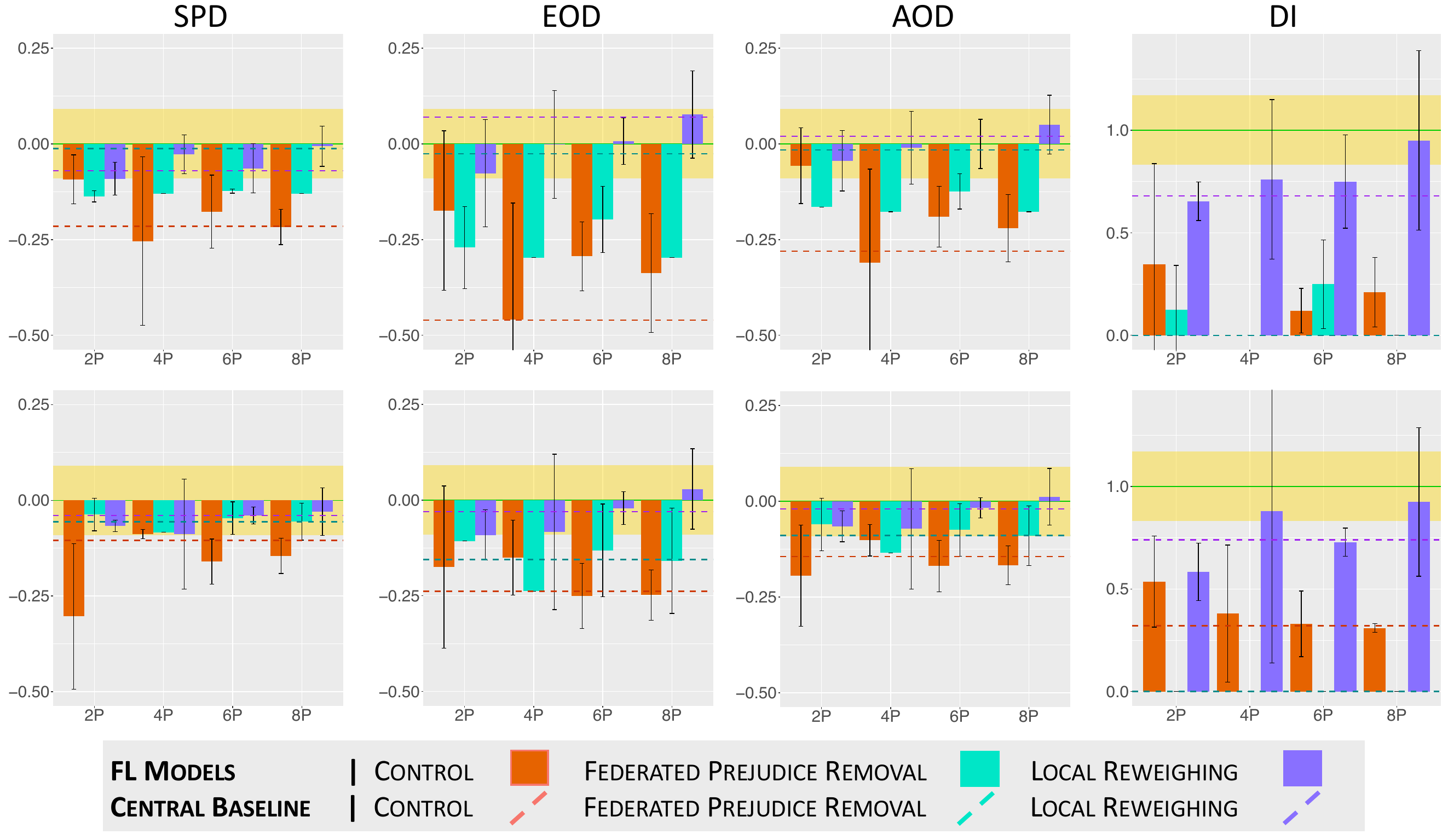}
\caption{\footnotesize Fairness metrics of models trained on stratified \textit{Adult} dataset against \textit{sex} (first row) and \textit{race} (second row) attributes}
\label{graph:adult-stratified-fairness}
\end{figure}

\begin{figure}[h]
\includegraphics[width=\columnwidth]{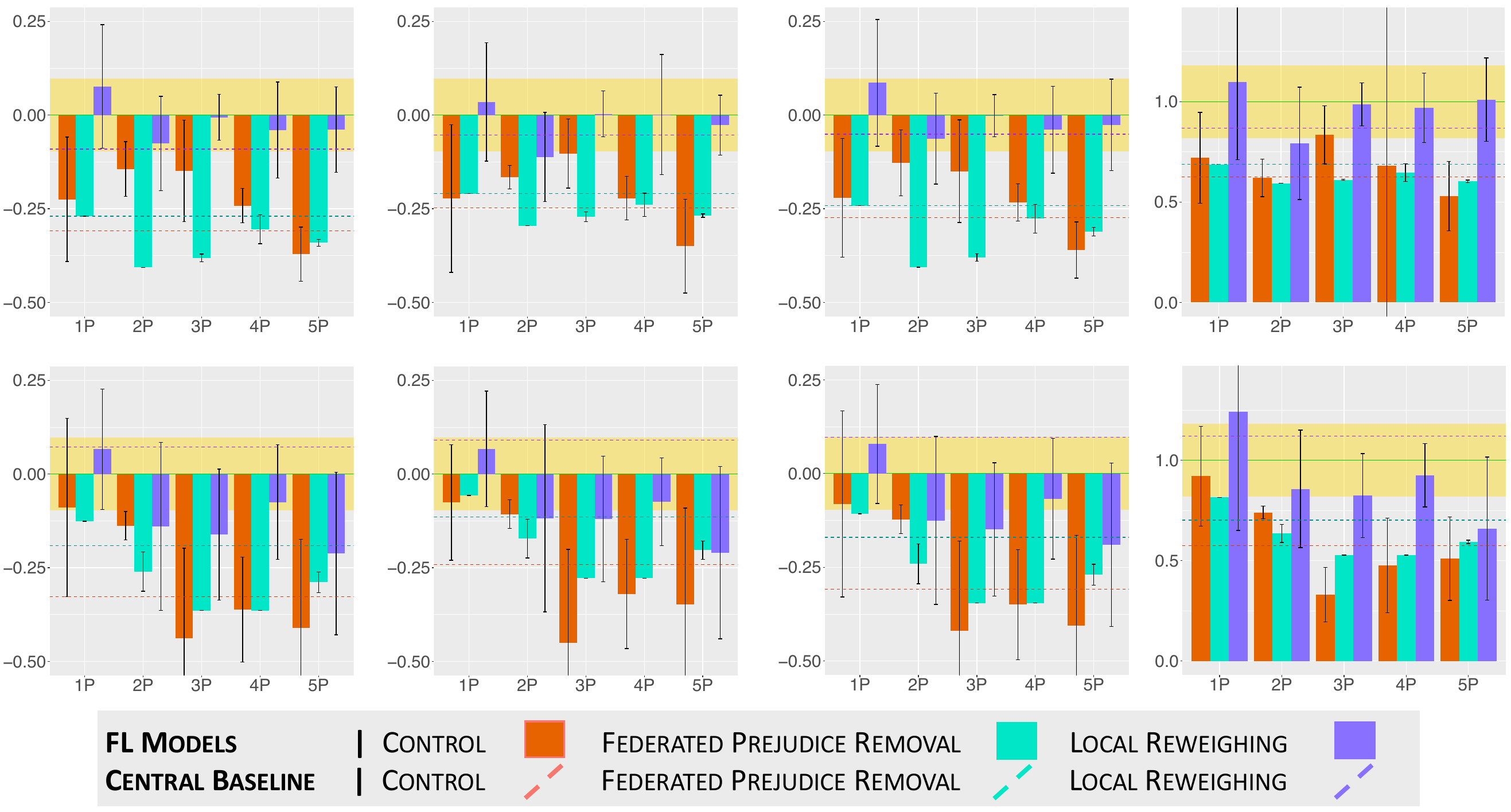}
\caption{\footnotesize Fairness metrics of models trained on stratified \textit{Compas} dataset against \textit{sex} (first row) and \textit{race} (second row) attributes}
\label{graph:compas-stratified-fairness}
\end{figure}

\begin{figure*}[h]
\centering
\includegraphics[width=0.7\columnwidth]{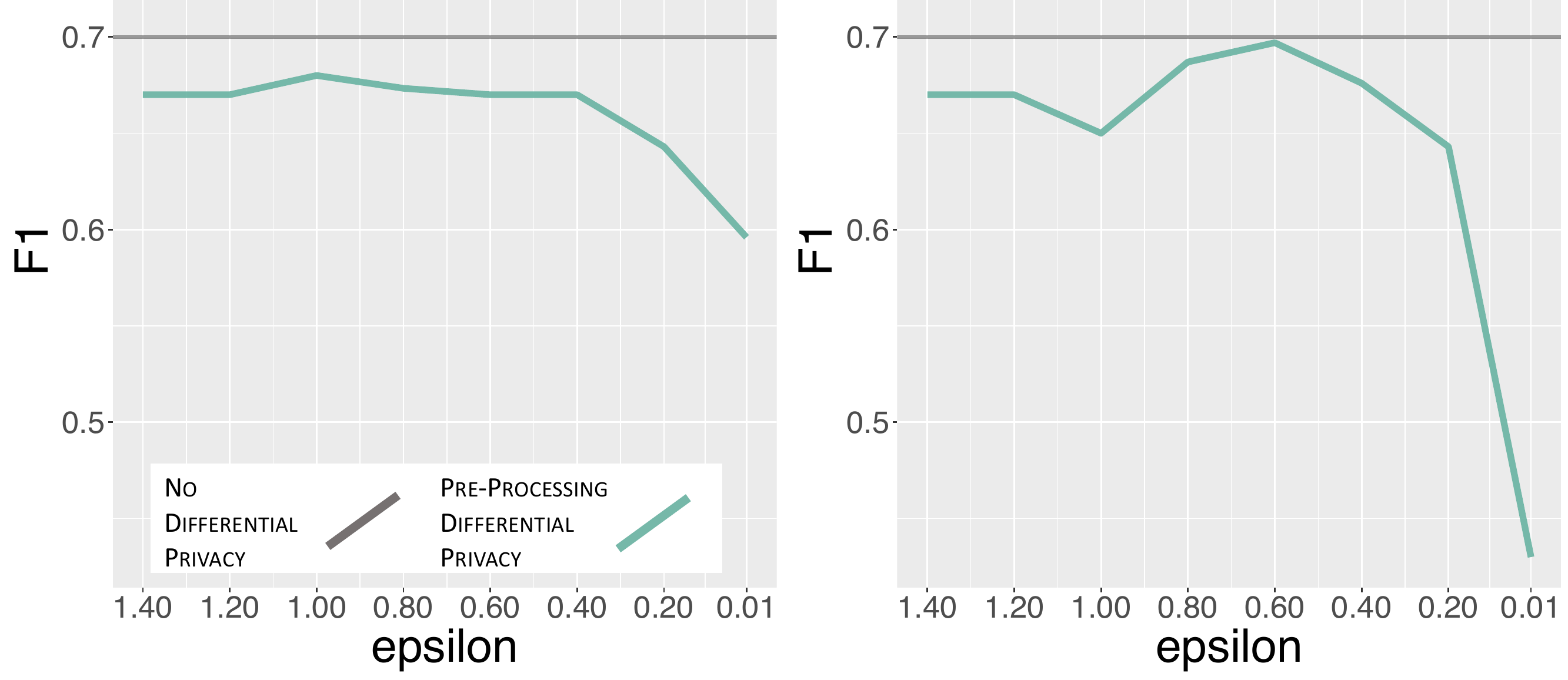}
\caption{\footnotesize Effect of pre-processing privacy budgets on F1 scores; 8-party model trained on \textit{Adult} dataset against \textit{sex}(left) and \textit{race} (right) attributes}
\label{graph:epsilon-F1}
\end{figure*}

We report the \textit{local reweighing} and \textit{federated prejudice removal} experiment results in Figure \ref{graph:adult-stratified-fairness} for fairness performance and Figure \ref{graph:adult-stratified} for prediction performance, respectively, with \textit{global reweighing with differential privacy} results in figure \ref{graph:epsilon-F1}.

For figures \ref{graph:adult-stratified-fairness} and \ref{graph:adult-stratified}, the control group (in orange) denotes models trained in FL without bias mitigation.
On the x-axis are the number of parties in the FL system, and the value of the corresponding metric is presented on the y-axis. 
The dashed lines illustrate the centralized baseline values, and the shaded region represents an acceptable margin of fairness metrics, as defined in \cite{bellamy2018ai}.

On one hand, we observe from Figure \ref{graph:adult-stratified-fairness} that for both the \textit{sex} and \textit{race} attributes, \textit{local reweighing} not only reduces bias, but also mitigates it such that it is within the acceptable margin for all fairness metrics. 
In addition, the resulting accuracy and F1 scores are consistent post-local reweighing.

We can conclude that \textit{local reweighing} is effective in mitigating bias for FL with IID party data. On the other hand, \textit{federated prejudice removal} with $\eta_{sex}$ = 1.25 and $\eta_{race}$ = 11.5 reduces bias in 3 out of 4 metrics for both sensitive attributes,
although not within the fairness margin. 
Moreover, from Figure \ref{graph:adult-stratified} we notice a trend between the fairness and performance metrics, where good performance of the fairness metrics is often coupled with lower accuracy and F1 scores. 
Contradicting the observations from centralized ML, this trend is particularly evident in \textit{federated prejudice removal} experiments, where increasing $\eta$ sharply drops F1 scores. Models trained with \textit{federated prejudice removal} perform in a much more stable manner than those trained with \textit{local reweighing} for all metrics.

We report the \textit{Compas} experiment results in Figures \ref{graph:compas-stratified-fairness} and \ref{graph:compas-acc-stratified} for fairness performance and prediction performance, respectively. 
On one hand, we observe from Figure \ref{graph:compas-stratified-fairness} that for both the \textit{sex} and \textit{race} attributes, \textit{local reweighing} not only reduces bias, but also mitigates it such that it is within the acceptable margin for a majority of fairness metrics. In addition, the resulting accuracy and F1 scores are consistent post-local reweighing. This supports the conclusion in the main text that \textit{local reweighing} is effective in mitigating bias for FL with IID party data. 
On the other hand, \textit{federated prejudice removal} with $\eta_{sex}$ = 1.5 and $\eta_{race}$ = 0.75 reduces bias in half of metrics for the sensitive attributes,
although not within the fairness margin. We again notice a trend, this time in Figure \ref{graph:compas-acc-stratified} between the fairness and performance metrics, where good performance of the fairness metrics is often coupled with lower accuracy and F1 scores. Models trained with \textit{federated prejudice removal} perform in a much more stable manner than those trained with \textit{local reweighing} for all metrics.

Similar to \textit{Adult} dataset, experiments results on \textit{Compas} also demonstrate several key factors that play a role in affecting model fairness in FL.

We assess the performance of our \textit{global reweighing with DP} approach by looking at the 8-party case for the \textit{Adult} dataset. We examine how a range of privacy budgets utilized in the pre-processing stage affect both model and fairness performance. Figure \ref{graph:epsilon-F1} shows the trade-off between privacy and model performance.

Recall that a larger $\epsilon$ results in less noise injection and hence less privacy.

The results show that the F1 scores decrease slightly as $\epsilon$ decreases, but is overall consistent until $\epsilon$ is very small.

\begin{table}[h]
\caption{ Fairness metrics when choosing different $\epsilon$, 8-party model trained on \textit{Adult}  against \textit{sex} attribute}
\label{fairness-epsilon}
\begin{center}
\begin{small}
\begin{sc}
\begin{tabular}{ccccc}
\toprule
$\epsilon$ & SPD & EOD & AOD & DI\\
\midrule
1.4 & $-0.05 \pm 0.01$  & $0.01 \pm 0.01$ & $0.01 \pm 0.01$ & $\mathbf{0.71} \pm 0.01$\\

1.0 & $-0.04 \pm 0.01$ & $-0.02 \pm 0.09$ & $0.00 \pm 0.06$ & $0.65 \pm 1.11$\\

0.8 & $-0.06 \pm 0.01$ & $0.00 \pm 0.02$ & $0.00  \pm 0.01$ & $0.71 \pm 0.05$\\

0.4 & $0.02 \pm 0.09$ & $\mathbf{0.14} \pm 0.26$ & $0.09 \pm 0.16$ & $1.84 \pm 1.78$\\

0.2 & $-0.05 \pm 0.05$ & $0.04 \pm 0.04$ & $0.02 \pm 0.04$ & $0.85 \pm 0.05$\\

0.01 & $0.05 \pm 0.07$ & $\mathbf{0.20} \pm 0.17$ & $\mathbf{0.13} \pm 0.11$ & $\mathbf{2.02} \pm 1.62$\\
\bottomrule
\end{tabular}
\end{sc}
\end{small}
\end{center}
\vskip -0.2in
\end{table}

We also measure the effect of privacy budgets on the model's fairness metrics shown in Table \ref{fairness-epsilon}.
Metrics that do not fall within the predefined fairness bounds are highlighted in red. The vast majority of fairness metrics are considered fair until $\epsilon = 0.01$. In Figure \ref{graph:epsilon-F1}, the F1 score is consistent, then drops significantly after $\epsilon$ is below 0.4. Pairing these results, we conclude that \textit{global reweighing with DP} is effective with $\epsilon$ values as low as $0.4$ for models trained on the \textit{Adult} dataset.

We now analyze several additional key factors that may play a role in affecting model fairness in FL. 

\noindent \textit{Number of parties:}
We train our FL models while increasing the number of parties, and hence with more data samples indirectly ``contributing'' to the final FL model. 
Note that under the IID data setting, we largely remove any other influence in FL aside from the number of parties. From Fig.\ref{graph:adult-stratified} and \ref{graph:adult-stratified-fairness}, we do not see that FL models trained with more parties achieve better fairness, while F1 scores do show improvement.

\begin{figure}[h]
\includegraphics[width=\columnwidth]{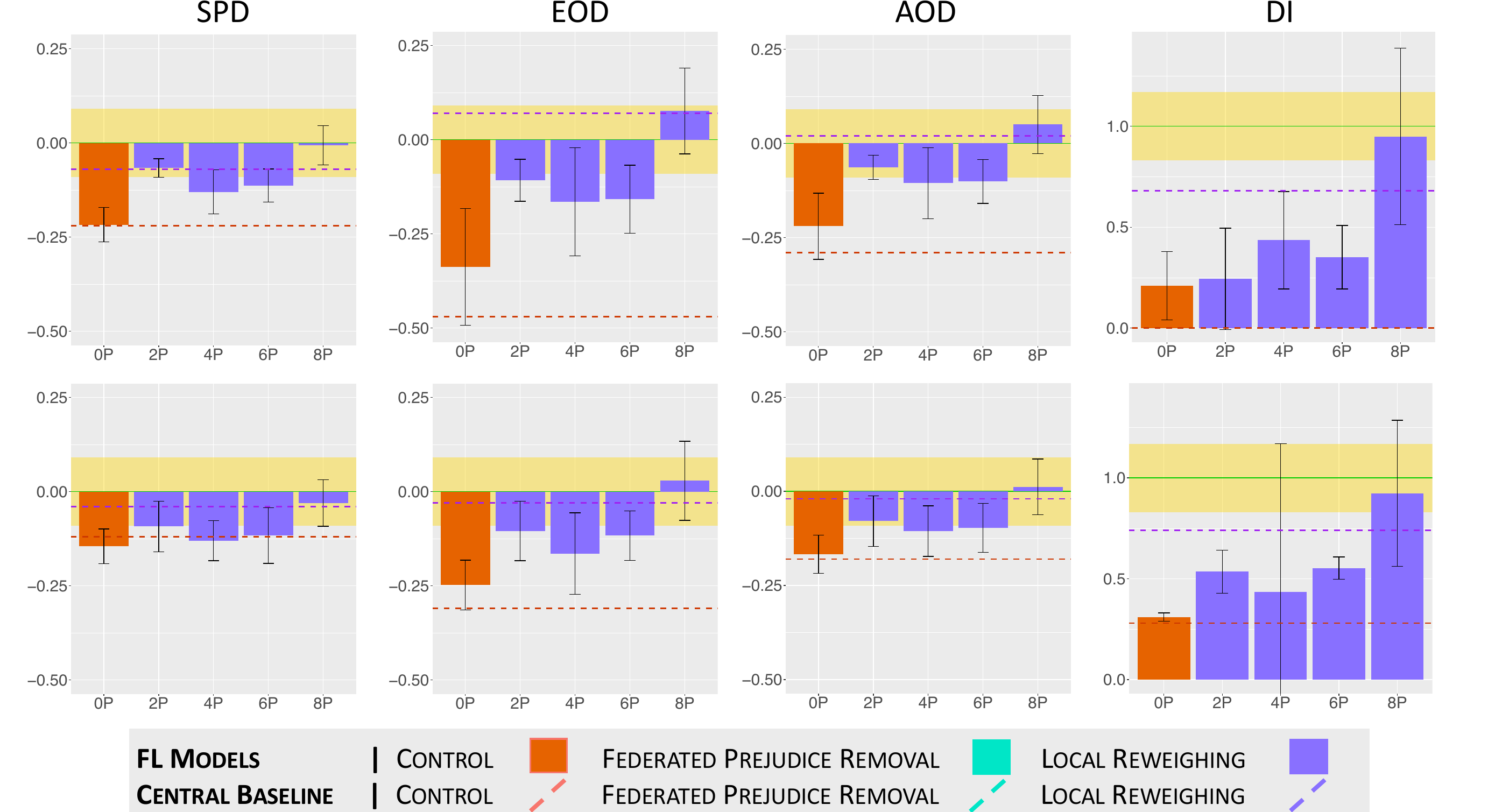}
\caption{\footnotesize Fairness metrics of FL models with partial \textit{local reweighing} on stratified \textit{Adult} dataset against \textit{sec} (first row) and \textit{race} (second row) attribute}
\label{graph:adult-stratified-fairness-prw-test}
\end{figure}

\begin{figure}[h]
\centering
\includegraphics[width=\columnwidth]{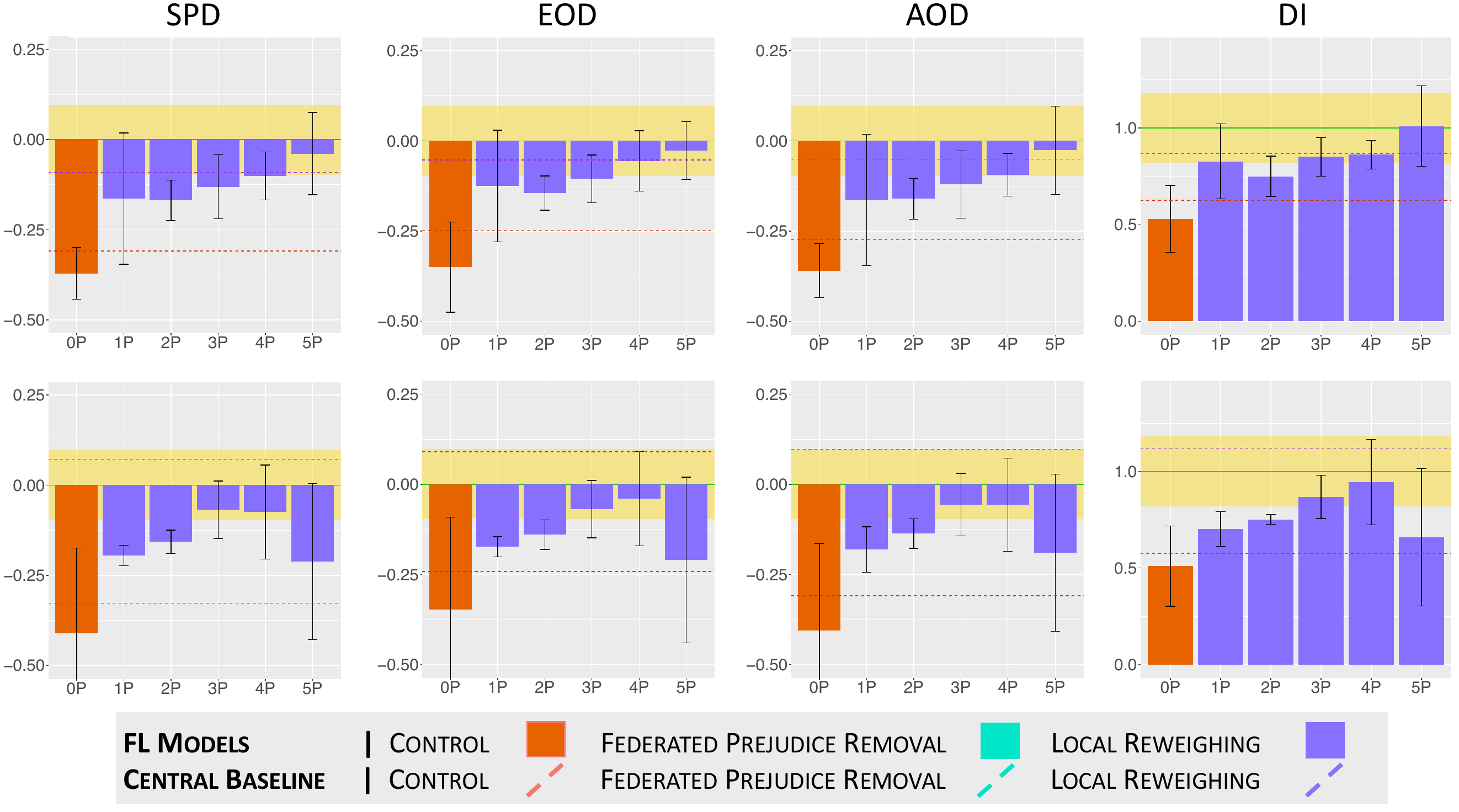}
\caption{\footnotesize Fairness metrics of FL models with partial \textit{local reweighing} on stratified \textit{Compas} dataset against \textit{sec} (first row) and \textit{race} (second row) attribute}
\label{graph:compas-prw}
\end{figure}

\noindent \textit{Partial local reweighing:}
We test the effects of having only a percentage of the parties,  i.e. 25\%, 50\%, 75\% and 100\% of eight parties, apply \textit{local reweighing} in FL. For the Compas dataset, 20\%, 40\%, 60\% , 80\% and 100\% of parties.
From Figure~\ref{graph:adult-stratified-fairness-prw-test}, we see that even when the lowest ratio utilizes \textit{local reweighing}, we see a large effect on the amount of bias detected in the global model, indicating that A) \textit{local reweighing} does not require all parties to participate to be effective, and with that B) the decision to perform pre-processing locally serves the users better. 
The latter part is very important for FL, as parties may join and drop out during the training process, and consequently, enforcing them to consistently perform pre-processing techniques is difficult. We ALSO test the effects of having only a percentage of the parties,  i.e. 20\%, 40\%, 60\%, 80\% and 100\% of five parties, apply \textit{local reweighing} in FL on the Compas dataset. From Figure~\ref{graph:compas-prw}, we see that even when the lowest ratio utilizes \textit{local reweighing}, we see a large effect on the amount of bias detected in the global model, which is in line with our \textit{Adult} dataset results.

\noindent \textit{Comparison with baseline:}
Both the baseline model and the FL models with eight parties are trained over the entirety of the training set, the difference being that the baseline is trained in a centralized fashion and the FL models are trained without parties sharing the training data.
As seen in Figure \ref{graph:adult-stratified-fairness}, the FL models with bias mitigation techniques scored fairer than the baseline in 3 out of 4 metrics across both sensitive attributes.
Moreover, the FL models without bias mitigation, i.e., the control group in orange, detect equal or less bias than the orange baseline across both sensitive attributes. 
This indicates that the model learns less bias simply by being trained in a FL setting with IID data.

\subsection{Performance under highly imbalanced data distributions (non-IID)}\label{sec:exp-imbalanced}
This subsection is devoted to examining FL settings that are likely to happen in the real world, where parties' data distribution is imbalanced. 
Individual parties could attempt to evaluate models with respect to a sensitive attribute and have very few or zero samples from either the unprivileged or privileged group, which may affect bias mitigation.
To simulate this scenario, we create two parties of $3,735$ samples each from \textit{Adult} dataset, but adjust their ratio of privileged group to unprivileged group. 
Experiments are run with 85-15, 99-1 and 100-0 ratio settings, e.g., regarding \textit{sex} attribute, Party 1 has 85\% male and 15\% female samples, and Party 2 has 85\% female and 15\% male samples.

\begin{figure*}[h]
\centering
\includegraphics[width=0.7\columnwidth]{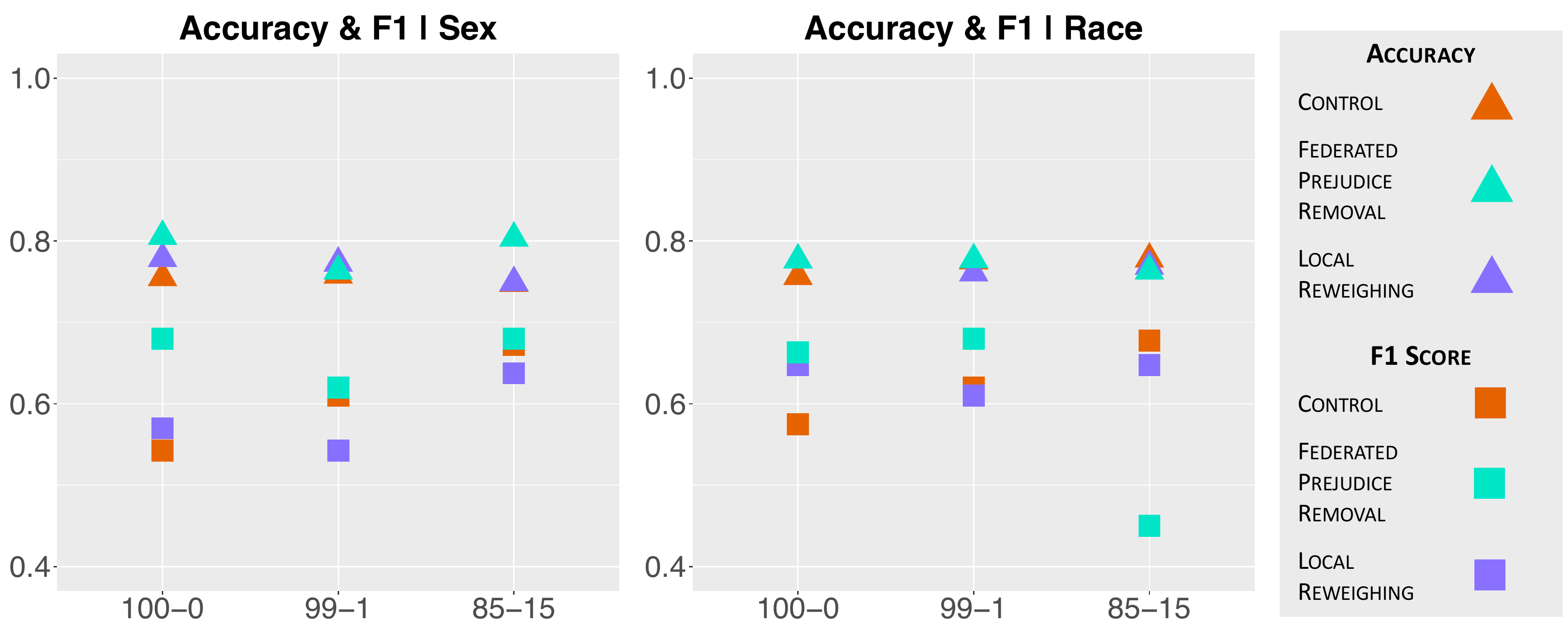}
\caption{\footnotesize Performance metrics of models trained on highly imbalanced \textit{Adult} dataset}
\label{graph:imbalanced-acc}
\end{figure*}

\begin{figure}[h]
\centering
\includegraphics[width=\columnwidth]{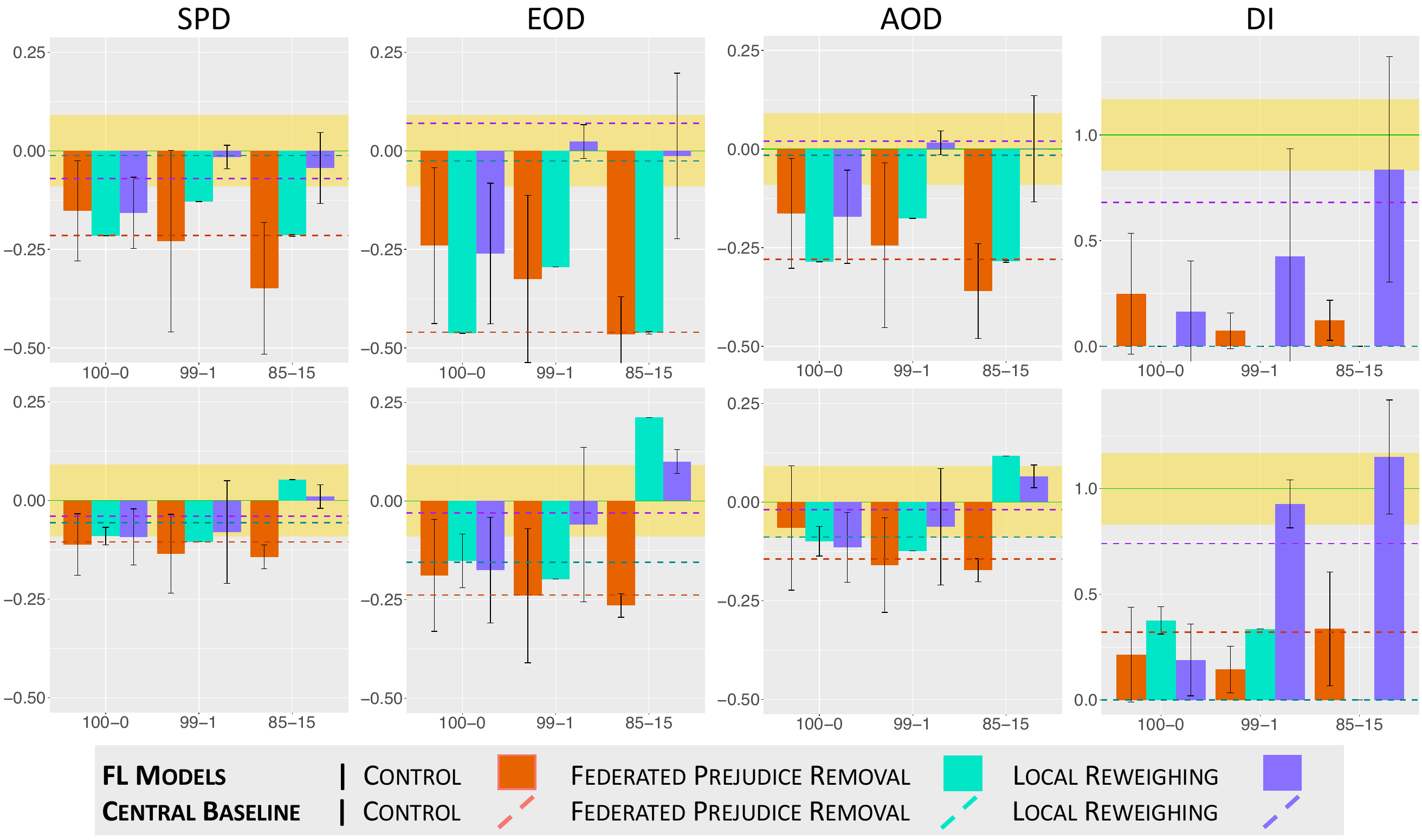}
\caption{\footnotesize Fairness metrics of models trained on highly imbalanced \textit{Adult} dataset against \textit{sex} (first row) and \textit{race} (second row) attribute}
\label{graph:exp-imbalanced-fairness}
\end{figure}

Figure \ref{graph:exp-imbalanced-fairness} and \ref{graph:imbalanced-acc} display the fairness and performance metric results from the highly imbalanced experiments, respectively. On the x-axis are the unprivileged-privileged ratios in either party, and the value of the corresponding metric is presented on the y-axis. 
From Figure~\ref{graph:exp-imbalanced-fairness}, we see \textit{local reweighing} has little to no effect on mitigating bias in the 100-0 proportion case.

It is reasonable, since if a party has samples from only the privileged or the unprivileged group, $W(s,y)$ is always 1 according to \eqref{eq:reweigh}.

Therefore, \textit{local reweighing} is ineffective for parties with disjoint sample sets. 
However, as long as a party has samples from both privileged and unprivileged groups, as low as $1\%$ of either group, \textit{local reweighing} can be effective for bias mitigation, and in Figure~\ref{graph:exp-imbalanced-fairness} most fairness metrics are within the fairness margin.
As parties' data distribution becomes more balanced, we see both the model's prediction and fairness performance improve. 
On the other hand, for both sensitive attributes, \textit{federated prejudice removal}'s efforts, with $\eta_{sex}$ = 1.25 and $\eta_{race}$ = 11.5, vary in reducing bias. 
Though it is effective to mitigate some bias compared to the control group and the baseline, it is less effective than \textit{local reweighing}.
Moreover, the converse relationship between the fairness and performance metrics is evident as shown in Figure \ref{graph:imbalanced-acc}.
In particularly, F1 scores drop significantly as models becomes fairer.

\begin{figure}[h]
\includegraphics[width=\columnwidth]{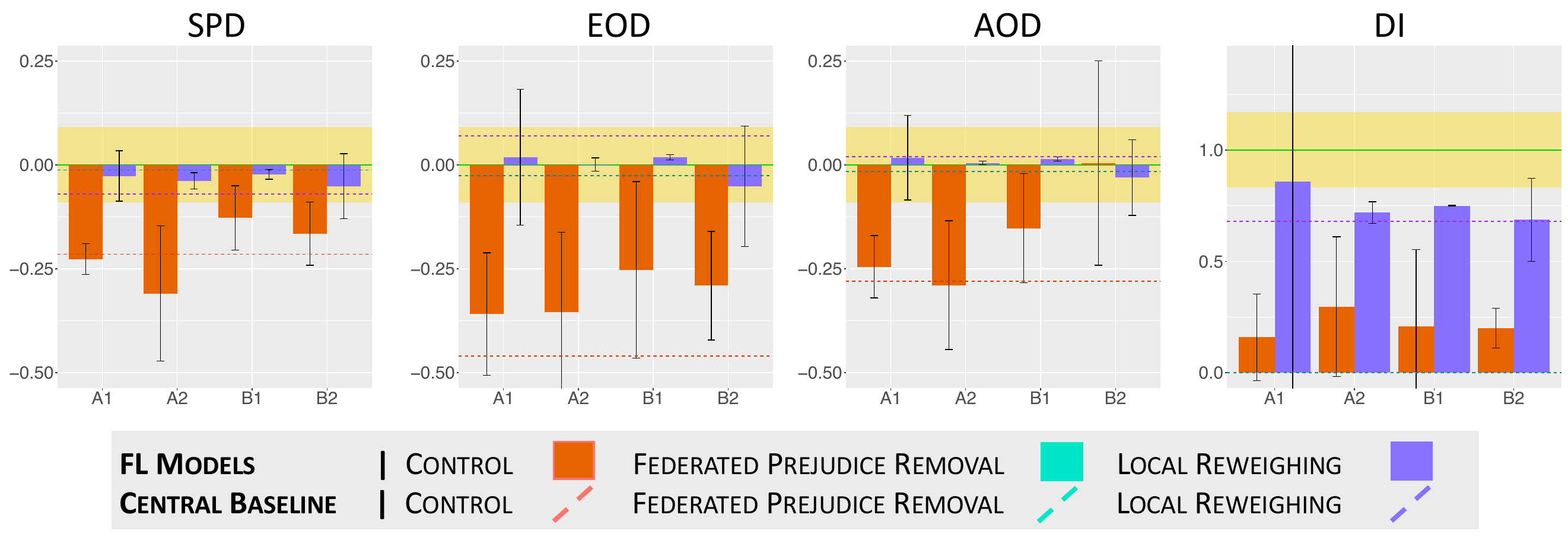}
\caption{\footnotesize Fairness metrics of models trained on stratified \textit{Adult} dataset against \textit{sex} attribute}
\label{graph:adult_5P_diff_size}
\end{figure}

Next we assess the performance of \textit{local reweighing} for scenarios where parties have 
both highly imbalanced data distributions and different sample sizes.
We examine \textit{local reweighing} for FL systems with five parties having different dataset conditions, as detailed in Table \ref{party-priv-tb}.
Groups A (containing A1 and A2) and B (containing B1 and B2) are similar in unprivileged:privileged ratio, but the majority group between underprivileged and privileged is flipped, e.g., party 4 is 80\% unprivileged and 20\% privileged in group A, and 20\% unprivileged and 80\% privileged in group B. Within the group, we have cases 1 and 2, where the cases have the same unprivileged/privileged ratios, but differ in that Case 1 has parties of all equal sizes, and Case 2 has unequal sizes. 
As displayed in Figure \ref{graph:adult_5P_diff_size}, different highly imbalanced data distributions, different party sizes, and the combination of the two do not hinder the effectiveness of \textit{local reweighing} as a bias mitigation technique.

\begin{table}[ht]
\caption{\footnotesize Sample ratio of unprivileged to privileged. Total samples shown in parentheses}
\label{party-priv-tb}
\centering
\begin{small}
\begin{sc}
\resizebox{0.75\columnwidth}{!}{%
\begin{tabular}{ccccc}
\toprule
Party & A1 & A2 & B1 & B2\\
\midrule
1 & 50:50 (2000) & 50:50 (500)& 50:50 (2000)& 50:50 (500)\\
2 & 50:50 (2000)& 50:50 (1500)& 50:50 (2000)& 50:50  (1500)\\
3 & 80:20 (2000)& 80:20 (2000)& 20:80 (2000)& 20:80  (2000)\\
4 & 90:10 (2000)& 90:10 (800)& 10:90 (2000)& 10:90 (800)\\
5 & 60:40 (2000)& 60:40 (1700)& 40:60 (2000) & 40:60 (1700)\\
\bottomrule
\end{tabular}
}
\end{sc}
\end{small}
\vskip -0.1in
\end{table}

\begin{figure}[h]
\centering
\includegraphics[width=0.7\columnwidth]{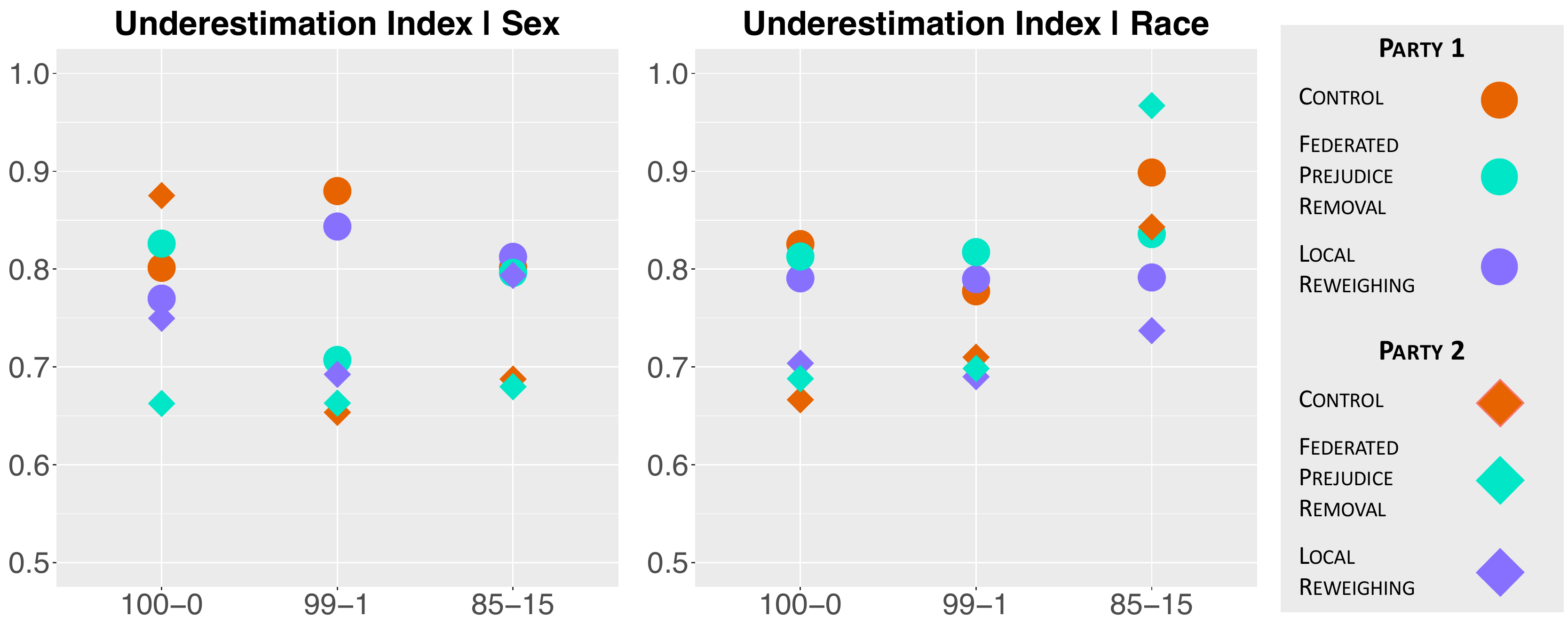}
\caption{\footnotesize UEI of models trained on highly imbalanced \textit{Adult} against \textit{sex} and \textit{race} attribute}
\label{graph:uei}
\end{figure}

\noindent \textit{Uneven bias learning:}
We find that FL models trained over highly imbalanced data distributions may learn bias from parties in an imbalanced manner. 
We utilize an \textit{underestimation index (UEI)} \cite{kamishima2012}, defined as the \textit{Hellinger distance} between the training data-induced distribution and the trained model-induced distribution, to measure the similarity between the truth and model predictions.
Lower UEI indicates greater similarity.  
As shown in Figure \ref{graph:uei}, the UEI of models trained via FL has significantly higher values for Party 1 than those for Party 2 across most ratio combinations, indicating that the global model consistently leans towards Party 2's training dataset.
This is underscored in the 100-0 case, as we find that the amount of bias in the FL model is roughly equivalent to that detected in Party 2's local model.

\begin{figure}[h]
\centering
\includegraphics[width=\columnwidth]{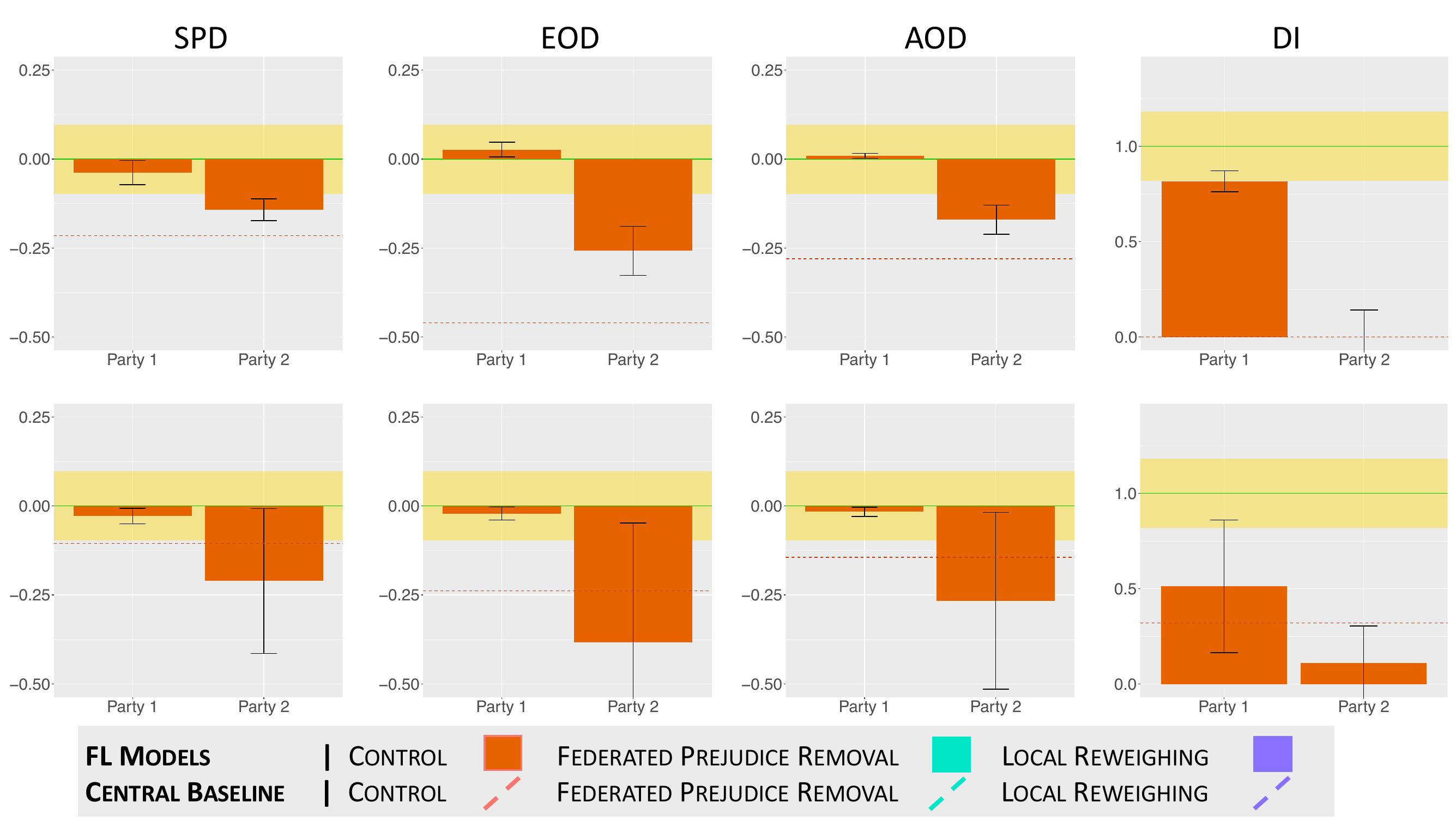}
\caption{\footnotesize Fairness metrics of local models trained on highly imbalanced \textit{Adult} dataset}
\label{graph:localmodels}
\end{figure}

In Figure \ref{graph:localmodels}, we display the fairness metrics of both parties' local models without any bias mitigation. Across the majority of metrics, Party 1's local model scores within the fairness margin, and significantly fairer than Party 2's local model for all metrics. The amount of bias detected in Party 2 is approximately equal to the amount of bias detected in the global 100-0 model, which falls in line with our analysis of the underestimation index values between the two parties; this supports the claim that federated models trained over parties with high data heterogeneity may learn their bias in an imbalanced way.

These results support the claim that fusion algorithms may induce bias by the way model updates are averaged, as discussed in Section \ref{sec:causes}.

\noindent \textit{Stability of fairness metrics:}
As all experiments are replicated three times, we expect consistent metric results. However, observed from Figures \ref{graph:adult-stratified-fairness},  \ref{graph:adult-stratified-fairness-prw-test} and \ref{graph:exp-imbalanced-fairness}, DI has large standard deviations (STD) across all FL experiments; for example, DI has a mean of $0.84$ (within the acceptable margin) and $0.55$ STD for the highly imbalanced, 85-15 \textit{sex} attribute case, while for the centralized baseline DI has 0 STD.
With a fairness margin width of 0.4, a STD of nearly 1.5 times that makes most DI values unreliable.
In FL, the other three metrics have an approximate STD of $0.2$, which seems more stable, and hence reliable. Amongst all fairness metrics discussed, DI is the only metric calculated as the ratio of the favorable outcome between the underprivileged and privileged group, and varies drastically as the model flips a few predictions.

Across the board, \textit{local reweighing} is a more effective method to drastically reduce FL model bias, even in spite of reduced mitigation participation or disproportionate data. Additionally, as shown in Table \ref{tab:comparison}, no communication rounds are added, no hyperparameters are needed and data private is preserved.

\section{Conclusion}\label{sec:end}
Issues of detecting and correcting undesired bias in FL systems have not been addressed yet. Parties’ data sets may be heterogeneous and the participation may change over the course of the training process.
To the best of our knowledge, this is the first paper to systematically assess these questions, against commonly used bias metrics, and in FL settings.
We have presented and contrasted three different methods to ensure models produced in FL are fair, and have demonstrated that some metrics lack stability.
We have also utilized UEI, which according to our experiments provides a better picture of how a global model is influenced by parties' local datasets. 
Our results demonstrate that utilizing \textit{local reweighing} produces fair models without sacrificing privacy or model accuracy, even when only a small fraction of parties engage in the fairness procedure. We hope this paper inspires further research in this area.

\newpage
\bibliographystyle{unsrt}

\end{document}